\documentclass{llncs}

\usepackage{epsfig}
\usepackage{latexsym}
\usepackage{amsfonts}
\usepackage{amsmath}
\usepackage{url}

\usepackage{makeidx}  % allows for indexgeneration
\begin{document}
\frontmatter          % for the preliminaries
\pagestyle{headings}  % switches on printing of running heads

\newcommand{\argmin}{\mathrm{argmin}}
\newcommand{\rk}{\mathrm{rk}}
\newcommand{\nd}{\mathrm{nd}}

\title{Median topographic maps for biomedical data sets}
\author{Barbara Hammer\inst{1} \and Alexander Hasenfuss\inst{1} \and
Fabrice Rossi\inst{2}}
\authorrunning{Hammer, Hasenfuss, Rossi} 
\institute{Clausthal University of Technology, D-38678 Clausthal-Zellerfeld, Germany,
\and
INRIA Rocquencourt, Domaine de Voluceau, Rocquencourt, B.P. 105,
78153 Le Chesnay Cedex, France}

\maketitle              % typeset the title of the contribution

\begin{abstract}
Median clustering extends popular neural data analysis methods such as
the self-organizing map or neural gas to general data structures given
by a dissimilarity matrix only. This offers flexible and robust
global data inspection methods which are particularly suited
for a variety of data as occurs in biomedical domains.
In this chapter, we give an overview about median clustering and its
properties and extensions, with a particular focus on efficient
implementations adapted to large scale data analysis.
\end{abstract}
\section{Introduction}
The tremendous growth of electronic information in biological and medical domains
has turned automatic data analysis and data inspection tools towards
a key technology for many application scenarios.
Clustering and data visualization constitute one fundamental problem
to arrange data in a way understandable by humans.
In biomedical domains, prototype based methods are particularly 
well suited since they represent data
in terms of typical values which can be directly inspected
by humans and visualized in the plane 
if an additional low-dimensional
neighborhood or embedding is present.
Popular methodologies include K-means clustering,
the self-organizing map, neural gas, affinity propagation, etc.
which have successfully been applied to various problems in
the biomedical domain such as
gene expression analysis, inspection of mass spectrometric data, health-care, analysis of microarray 
data, protein sequences, medical image analysis,
etc.\ \cite{alharbi,kaski,kaskibmc,mediansom,lu,shamir,villyannpr}. 

Many popular prototype-based clustering algorithms, however,
have been derived for
Euclidean data embedded in a real-vector space.
In biomedical applications,
data are diverse including
temporal signals such as EEG and EKG signals, 
functional data such as mass spectra, sequential
data such as DNA sequences, complex graph structures such as
biological networks, etc.
Often, the Euclidean metric is not appropriate to
compare such data, rather, a problem dependent similarity or
dissimilarity measure should be used such as 
alignment, correlation, graph distances, functional metrics, 
or general kernels.

Various extensions of prototype-based methods towards more
general data structures exist such as
extensions for recurrent and recursive data structures,
functional versions, or kernelized formulations,
see e.g.\ \cite{esann05hammertutorial,nn04hammer,nonstandard,barreto,ki2007hammer} for an overview.
A very general approach relies on 
a matrix which characterizes the pairwise similarities or
dissimilarities of data. This way, any 
distance measure or kernel (or generalization
thereof which might violate symmetry, triangle inequality, or
positive definiteness) can be dealt with including
discrete settings which cannot be embedded in Euclidean
space such as alignment of sequences 
or empirical measurements of pairwise similarities without
explicit underlying metric. 

Several approaches extend popular clustering
algorithms such as K-means or the self-organizing
map towards this setting by means of the relational dual
formulation or kernelization of the approaches
\cite{hathaway,hathaway2,kernelng,nathalie,ki2007hammer}.
These methods have the drawback that they partially require specific properties
of the dissimilarity matrix (such as positive definiteness),
and they represent data in terms of prototypes which are given
by (possibly implicit) mixtures of training points, thus they
cannot easily be interpreted directly. Another general approach leverages mean
field annealing techniques \cite{graepelsom,graepel,hofmann97pairwise} as a
way to optimize a modified criterion that does not rely anymore on the use of
prototypes. As for the relational and kernel approaches, the main drawback of
those solutions is the reduced interpretability. 

An alternative is offered by a representation of
classes by
the median or centroid, i.e.\
prototype locations are restricted to the discrete set
given by the training data.
This way, the distance of data points from prototypes
is well-defined.
The resulting learning problem is connected to a 
well-studied optimization problem,
the K-median problem:
given a set of data points and pairwise
dissimilarities, find $k$ points forming centroids and an assignment of the
data into k classes such that the average dissimilarities
of points to their respective closest centroid is minimized.
This problem is NP hard in general unless
the dissimilarities have a special form (e.g.\ tree metrics), 
and there exist constant factor approximations for specific settings
(e.g.\ metrics) \cite{charikar,arora}.

The popular K-medoid clustering extends the batch optimization
scheme of K-means to this restricted setting of prototypes: 
it in turn assigns data points to the respective closest
prototypes and determines optimum prototypes for these
assignments \cite{KaufmanRousseeuw1987PAM,CeleuxEtAl1989}.
Unlike K-means, there does not exist a closed form
of the optimum prototypes given fixed
assignments such that exhaustive search is used.
This results in a complexity ${\cal O}(N^2)$ for
one epoch for K-centers clustering instead of
${\cal O}(N)$ for K-means, $N$ being the number
of data points.
Like K-means, K-centers clustering is
highly sensitive to initialization.

Various approaches optimize the cost function
of K-means or K-median
by different methods to avoid local optima as much as
possible, such as Lagrange relaxations of the corresponding
integer linear program,
vertex substitution heuristics,
or affinity propagation \cite{vsh,affinity,affinity2}.
In the past years, simple, but powerful 
extensions of neural based clustering
to general dissimilarities have been proposed which
can be seen as generalizations of K-centers clustering
to include neighborhood cooperation, such
that the topology of data is taken into account.
More precisely, the median clustering
has been integrated into the popular self-organizing map (SOM) \cite{AmbroiseGovaert1996,KohonenSymbol1996}
and its applicability has been demonstrated in a large scale
experiment from bioinformatics \cite{mediansom}.
Later, the same idea has been integrated into
neural gas (NG) clustering together with
a proof of the convergence of
median SOM and median NG clustering \cite{marie}.
Like K-centers clustering, the methods
require an exhaustive search to obtain optimum prototypes
given fixed assignments such that the
complexity of a standard implementation for one epoch is ${\cal O}(N^2K)$
(this can be reduced to ${\cal O}(N^2+NK^2)$ for median SOM, see Section \ref{sectionFastMedianClustering} and \cite{FastDSOM}). 
Unlike K-means, local optima and overfitting
can widely be avoided
due to the neighborhood cooperation such that
fast and reliable methods result which are robust
with respect to noise in the data.
Apart from this numerical stability,
the methods have further benefits:
they are given by simple formulas and they are very easy to implement,
they rely on underlying cost functions which can be extended towards
the setting of partial supervision, and
in many  situations a considerable speed-up
of the algorithm can be obtained, as demonstrated in
\cite{FastDSOM,aipr}, for example.

In this chapter, we present an overview
about neural based median clustering.
We present the principle methods based on the cost
functions of NG and SOM, respectively, and 
discuss applications, extensions and properties.
Afterwards, we discuss several possibilities to speed-up
the clustering algorithms, including
exact methods, as well as single pass approximations for
large data sets.

\section{Prototype based clustering}
Prototype based clustering aims for representing
given data from some set $X$
faithfully by means of
prototypical representatives $\{\vec w^1,\ldots,\vec w^K\}$.
In the standard Euclidean case, real vectors are
dealt with, i.e.\ $X\subseteq\mathbb{R}^M$ and $\vec w^i\in\mathbb{R}^M$
holds for all $i$ and some dimensionality $M$.
For every data point $\vec x \in X$, the index of the \emph{winner} is defined as
the prototype
\begin{equation}
I(\vec x)=\argmin_j\{d(\vec x,\vec w^j)\}
\label{eq_winner}
\end{equation}
where
\begin{equation}
d(\vec x,\vec w^j)=\sum_{i=1}^M(x_i-w^j_i)^2
\label{eq_distance}
\end{equation}
denotes the squared Euclidean distance.
The \emph{receptive field} of prototype $\vec w^j$ is
defined as the set of data points for which it becomes winner.
Typically, clustering results are evaluated by
means of the \emph{quantization error} which
measures the distortion being introduced when data
is represented by a prototype, i.e.\
\begin{equation}
E:=\frac{1}{2}\cdot\int\sum_{j=1}^K\delta_{I(\vec x),j}\cdot d(\vec x,\vec w^j)P(\vec x)d \vec x 
\label{cont_eq_qerror}
\end{equation}
for a given probability measure $P$ according to
which data are distributed.
$\delta_{ij}$ denotes the Kronecker function.
In many training settings, 
a finite number of data
$X=\{\vec x^1,\ldots,\vec x^N\}$ is given in advance and the corresponding discrete
error becomes
\begin{equation}
\hat E:=\frac{1}{2N}\cdot\sum_{i=1}^N\sum_{j=1}^K\delta_{I(\vec x^i),j}\cdot d(\vec x^i,\vec w^j)
\label{eq_qerror}
\end{equation}

The popular K-means clustering algorithm aims at a
direct optimization of the quantization error.
In batch mode, it, in turn, determines
optimum prototypes $\vec w^j$ given fixed assignments $I(\vec x^i)$
and vice versa until convergence:
\begin{equation}
k_{ij}:=\delta_{I(\vec x^i),j},\quad \vec w^j:=\frac{\sum_ik_{ij}\vec x^i}{\sum_ik_{ij}}
\label{eq_kmeans}
\end{equation}
This update scheme is very sensitive to the initialization of prototypes
such that multiple restarts are usually necessary.

\subsection*{Neural gas}
The self-organizing map and neural gas enrich the update
scheme by neighborhood cooperation of
the prototypes. This accounts for a topological ordering
of the prototypes such that initialization sensitivity is (almost) avoided.
The cost function of neural gas as introduced by
Martinetz \cite{ng} has the form
\begin{equation}
E_{\mathrm{NG}}\sim\frac{1}{2}\cdot \int\sum_{j=1}^Kh_{\sigma}(\rk(\vec x,\vec w^j))\cdot d(\vec x,\vec w^j) P(\vec x)d\vec x
\label{cont_eq_ngerror}
\end{equation}
where
\begin{equation}
\rk(\vec x,\vec w^j)=\left|\left\{\vec w^k\:|\:d(\vec x,\vec w^k)<d(\vec x,\vec w^j)\right\}\right|
\label{eq_rank}
\end{equation}
denotes the rank of prototype $\vec w^j$ sorted according to its distance
from the data point $\vec x$ and
$h_{\sigma}(t)=\exp(-t/\sigma)$ is a Gaussian shaped curve
with the neighborhood range $\sigma>0$.
$\sigma$ is usually annealed during training.
Obviously, $\sigma\to0$ yields the standard
quantization error. For large values $\sigma$,
the cost function is smoothed such that
local optima are avoided at the beginning of training.
For a discrete training set, this cost term becomes
\begin{equation}
\hat E_{\mathrm{NG}}\sim\frac{1}{2N}\cdot \sum_{i=1}^N\sum_{j=1}^Kh_{\sigma}(\rk(\vec x^i,\vec w^j))\cdot d(\vec x^i,\vec w^j)
\label{eq_ngerror}
\end{equation}
This cost function is often optimized by means of a stochastic
gradient descent. An alternative is offered by batch clustering which,
in analogy to K-means, consecutively optimizes assignments
and prototype locations until convergence, as described in \cite{marie}:
\begin{equation}
k_{ij}:=\rk(\vec x^i,\vec w^j),\quad \vec w^j:=\frac{\sum_ih_{\sigma}(k_{ij})\vec x^i}{\sum_ih_{\sigma}(k_{ij})}
\label{eq_ng}
\end{equation}
Neighborhood cooperation takes place depending on the
given data at hand by means of the ranks.
This accounts for a very robust clustering scheme
which is very insensitive to local optima, as discussed
in \cite{ng}.
Further, as shown in \cite{trn}
neighborhood cooperation induces a topology on the
prototypes which perfectly fits the topology
of the underlying  data manifold provided the sampling is sufficiently dense.
Thus, browsing within this information space becomes possible.

\subsection*{Self-organizing map}
Unlike NG, SOM uses a priorly fixed lattice structure, often a regular
low-dimensional lattice such that
visualization of data can directly be achieved.
Original SOM as proposed by Kohonen \cite{kohonen}
does not possess a cost function in the continuous case,
but a simple variation as proposed by Heskes does \cite{heskes}.
The corresponding cost function
is given by
\begin{equation}
E_{\mathrm{SOM}}\sim \frac{1}{2}\cdot \int\sum_{j=1}^K
\delta_{I^*(\vec x),j}\cdot
\sum_{k=1}^N h_{\sigma}(\nd(j,l))\cdot d(\vec x,\vec w^l) P(\vec x)\vec x
\label{cont_eq_somerror}
\end{equation}
where 
$\nd(j,l)$ describes the distance
of neurons arranged on a priorly chosen
neighborhood structure of the
prototypes, often a regular two-dimensional lattice,
and
\begin{equation}
I^*(\vec x)=\argmin_i\left\{\sum_{l=1}^K h_{\sigma}(\nd(i,l)) d(\vec x,\vec w^l)\right\}
\end{equation}
describes the prototype which
is closest to $\vec x$ if averaged over the neighborhood.
(This is in practice often identical to the standard winner.)
In the discrete case, the cost function becomes
\begin{equation}
\hat E_{\mathrm{SOM}}\sim \frac{1}{2N}\cdot \sum_{i=1}^N\sum_{j=1}^K
\delta_{I^*(\vec x^i),j}\cdot
\sum_{k=1}^N h_{\sigma}(\nd(j,k))\cdot d(\vec x^i,\vec w^k)
\label{eq_somerror}
\end{equation}
SOM is often optimized by means of a stochastic
gradient descent or, alternatively, in a
fast batch mode, subsequently optimizing
assignments and prototypes as follows:
\begin{equation}
k_{ij}:=\delta_{I^*(\vec x^i),j},\quad
\vec w^j:=\frac{\sum_{i,k}k_{ik}h_{\sigma}(\nd(k,j))\vec x^i}{\sum_{i,k}k_{ik}h_{\sigma}(\nd(k,j))}
\label{eq_som}
\end{equation}
As before, the neighborhood range $\sigma$ is annealed to
$0$ during training, and the standard
quantization error is recovered.
Intermediate steps offer a smoothed cost function the optimization
of which is simpler such that local optima can widely be
avoided and excellent generalization
can be observed. Problems can occur if the priorly chosen topology
does not fit the underlying manifold and
topological mismatches can be observed.
Further, topological defomations can easily occur 
in batch optimization when
annealing the neighborhood quickly as demonstrated in \cite{mariebatchsom}.
These problems are not present for the data optimum topology provided
by NG, but, unlike SOM, NG does not offer a
direct visualization of data.
Batch SOM, NG, and K-means converge after a finite and
usually small number of epochs \cite{marie}.

\section{Median clustering}
Assume data is characterized only by a matrix of
pairwise nonnegative dissimilarities
\begin{equation}
D=(d(\vec x^i,\vec x^j))_{i,j=1,\ldots,N},
\label{matrix}
\end{equation}
i.e.\ it is not necessarily contained in an Euclidean space.
This setting covers several important
situations in biomedical domains such as
sequence data which are compared by alignment,
time dependent signals for which correlation analysis
gives a good dissimilarity measure, or medical
images for which problem specific dissimilarity measures 
give good results.
In particular, the data space $X$ is discrete such
that prototypes cannot be adapted smoothly within
the data space.

The idea of median clustering is to restrict prototype locations to data
points.
The objective of clustering as stated by the quantization
error (\ref{eq_qerror}) is
well defined if prototypes are restricted to the data set
$\vec w^j\in X$ and the
squared Euclidan metric is substituted by
a general term $d(\vec x^i,\vec w^j)$ given
by a dissimilarity matrix only. Similarly, the cost functions of
neural gas and the self-organizing map
remain well-defined for $\vec w^j\in X$
and arbitrary terms $d(\vec x^i,\vec w^j)$. Thereby, the
dissimilarities $D$ need not fulfill the conditions
of symmetry or the triangle inequality.

One can derive learning rules based on the cost functions
in the same way as batch clustering by means of a subsequent optimization
of prototype locations and assignments.
For NG, optimization of
the cost function
\[
\hat E_{\mathrm{NG}}\sim\frac{1}{2N}\cdot
\sum_{i=1}^N\sum_{j=1}^Kh_{\sigma}(\rk(\vec x^i,\vec w^j))\cdot d(\vec
x^i,\vec w^j)
\]
with the constraint $\vec w^j\in X$ yields the following algorithm
for median NG:
\begin{center}
\begin{tabular}{l}
init\\
repeat\\
\hspace*{1em} $k_{ij}:=\rk(\vec x^i,\vec w^j)$\\
\hspace*{1em} $\vec w^j:=\argmin_{\vec x^l}\sum_{i=1}^Nh_{\sigma}(k_{ij})d(\vec x^i,\vec x^l)$
\end{tabular}
\end{center}
Unlike batch NG, a closed solution for optimum prototype locations
does not exist and exhaustive search is necessary.
In consequence, one epoch has time complexity
${\cal O}(N^2K)$ compared
to ${\cal O}(NK)$ for batch NG (neglecting sorting of
prototypes).
Because of the discrete locations of
prototypes, the probability that prototypes
are assigned to the same location becomes nonzero.
This effect should be avoided e.g.\ by means of adding
random noise to the distances in every run or via an explicit
  collision prevention mechanism, as in \cite{Rossi2007EsannDSOMCollision} for the median SOM.

Similarly, median SOM can be derived from the cost
function
$\hat E_{\mathrm{SOM}}$.
The following algorithm is obtained:
\begin{center}
\begin{tabular}{l}
init\\
repeat\\
\hspace*{1em} $k_{ij}:=\delta_{I^*(\vec x^i),j}$\\
\hspace*{1em} $\vec w^j:=\argmin_{\vec x^l}\sum_{i=1}^N\sum_{k=1}^Kk_{ik}h_{\sigma}(\nd(k,j))d(\vec x^i,\vec x^l)$
\end{tabular}
\end{center}
As for median NG, prototypes have to be determined by
exhaustive search leading to ${\cal O}(N^2K)$ complexity per epoch.

\begin{figure}[tb]
\begin{center}
\includegraphics[width=12cm]{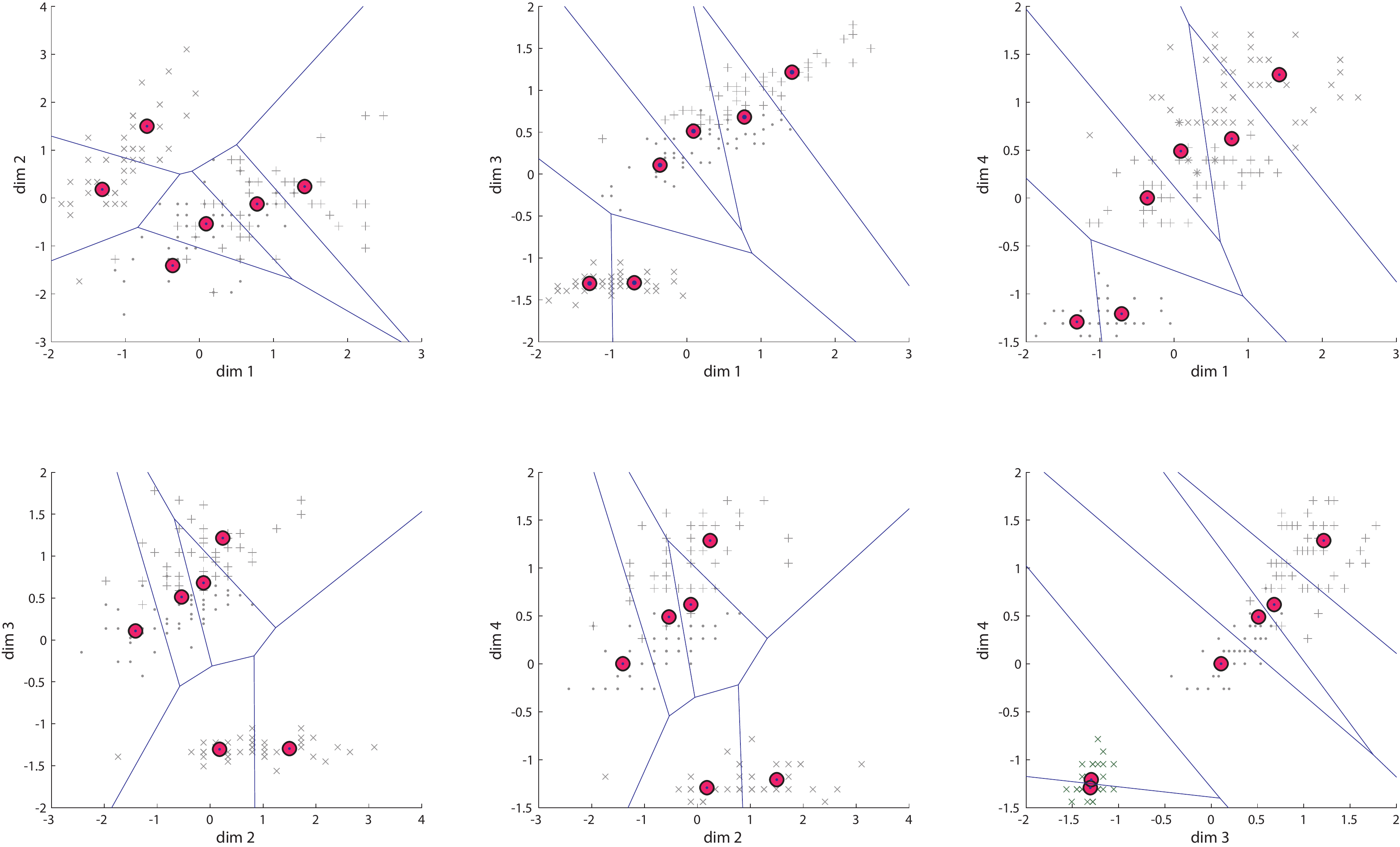}\\
\includegraphics[width=12cm]{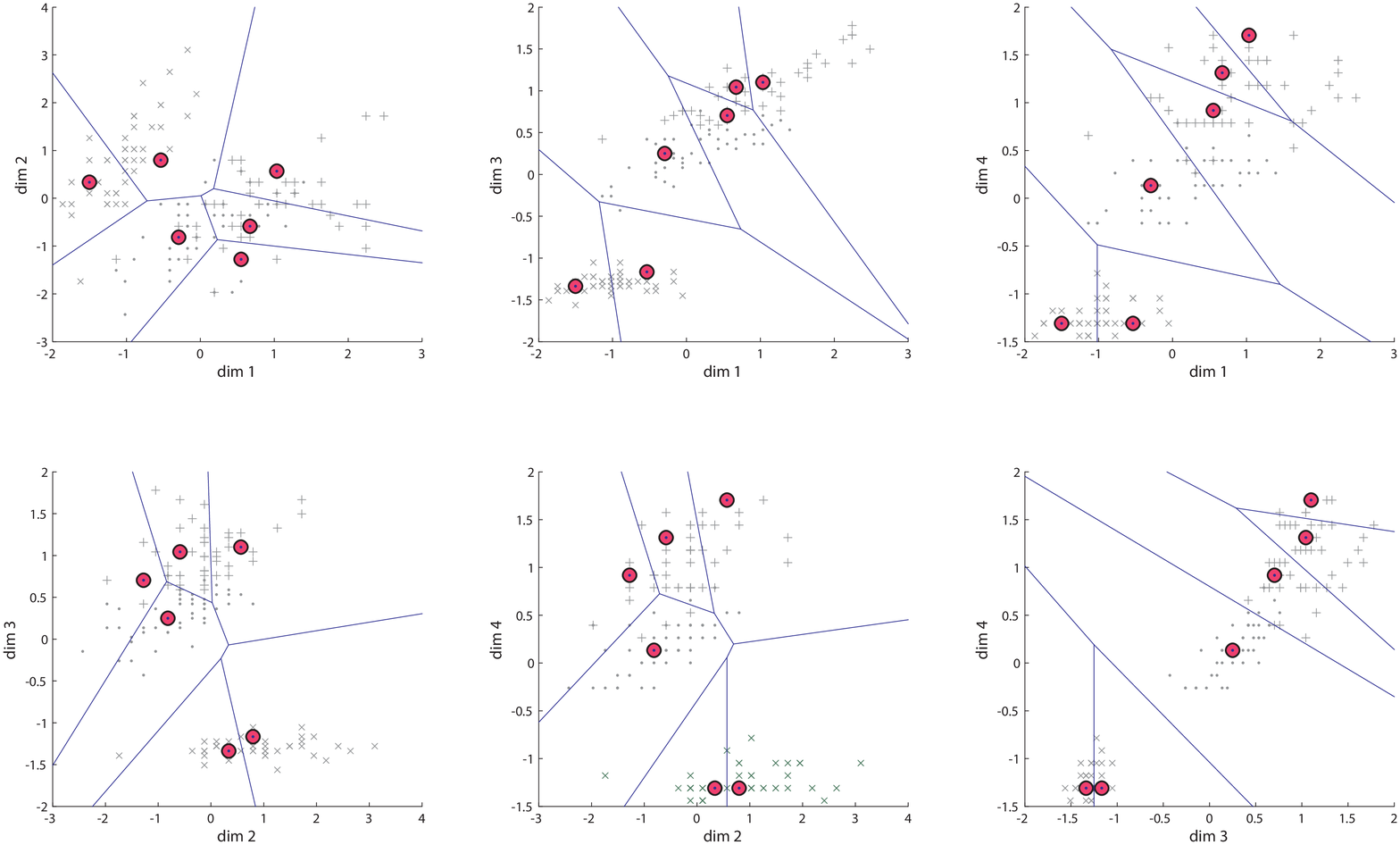}
\end{center}
\caption{Results of batch NG (top) and median NG (bottom)
on the iris data set projected to two of the four data dimensions.}
\label{fig_iris}
\end{figure}

Median clustering can be used for any given matrix
$D$. It has been shown in \cite{marie} that both, 
median SOM and median NG converge in a finite number of steps,
because both methods subsequently minimize the
respective underlying cost function until they arrive at a fixed
point of the algorithm. 
A simple demonstration of the behavior of median NG can be found
in Fig.~\ref{fig_iris}.
The popular iris data set (see \cite{Fisher,Anderson})
consists of 150 points and 3 classes.

Data are standardized to z-scores and batch NG and median NG
is applied using 6 prototypes.
Dimensions 3 and 4 are plotted
together with the prototypes found by batch NG and median NG, respectively
in Fig.~\ref{fig_iris}.
Obviously, the result is very similar, the quantization error
of batch NG being 40.96, while median NG displays the
slightly larger quantization error 44.85 due to
the restriction of prototypes to data points.
The classification accuracy obtained by posterior labeling
is 0.84 for batch NG as compared to 0.92 for median NG.
Since the prior class information is not taken
into account during training, standard batch NG yields a
larger classification error despite from its larger 
prototype flexibility.

\subsection*{Supervision}
Often, additional label information is available for (parts of) the
data set. Unsupervised data inspection and clustering
does not take this information into account and cluster boundaries
do not follow class distributions of the priorly known classes
unless they coincide with unsupervised clusters in the data.
This can be avoided by taking prior label information into account.
We assume that a class label $\vec y^i$ is available for 
every data point $\vec x^i$. We assume $\vec y^i\in\mathbb{R}^d$,
i.e.\ $d$ classes are represented by means of a full disjunctive
  coding (i.e., $y^i_j=\delta_{c^i,j}$, where $c^i$ is the index of the class
  for the data point $\vec x^i$),
including the possibility of fuzzy assignments.
We equip every prototype $\vec w^j$ with a label
$\vec Y^j\in\mathbb{R}^d$ which is adapted during training.
This vector represents the class label of the prototype, i.e.\
the average labels of data in its receptive field.

The aim of semisupervised clustering and
data inspection is to determine prototypes and their labels in such a way
that prototypes represent data point faithfully and they take
the labeling into account, i.e.\ the prototype labels should correspond to
the label of data points of its receptive field.
To achieve this goal, the distance of a prototype
$\vec w^j$ from a data point $\vec x^i$ is extended towards
\begin{equation}
d_{\beta}(\vec x^i,\vec w^j):=\beta\cdot d(\vec x^i,\vec w^j)+(1-\beta)\cdot d(\vec y^i,\vec Y^j)
\end{equation}
where $d(\vec y^i,\vec Y^j)$ denotes the squared Euclidean distance
of the labels $\vec y^i$ and $\vec Y^j$, and
$\beta\in(0,1)$ balances the goal to represent the input
data and the labels within the receptive field correctly.

This extended distance measure can be directly integrated into the
cost functions of NG and SOM.
Depending on the form of the distance measure, 
an extension of batch optimization or median optimization
becomes possible.
In both cases, the standard batch optimization scheme is accompanied by the label
updates, which yields
\begin{equation}
\vec Y^j=\sum_i h_{\sigma}(k_{ij})\cdot y^i/\sum_i h_{\sigma}(k_{ij})
\end{equation}
for batch and median NG, respectively, and
\begin{equation}
\vec Y^j=\sum_{ik} k_{ik}h_{\sigma}(\nd(k,j))\cdot y^i/\sum_{ik} k_{ik}h_{\sigma}(\nd(k,j))
\end{equation}
for batch and median SOM, respectively.
It can be shown in the same way as for standard batch and median
clustering, that these supervised variants converge after a finite number
of epochs towards a fixed point of the algorithm.

The effect of supervision can exemplarly be observed for the iris data set:
supervised batch NG with supervision parameter $\beta=0.5$
causes the prototypes to follow more closely the prior class borders
in particular in overlapping regions 
and, correspondingly,
an improved classification accuracy of
0.95 is obtained for supervised batch NG.
The mixture parameter $\beta$ constitutes a hyperparameter
of training which has to be optimized according to the
given data set. However, in general
the sensitivity with respect to $\beta$ seems to be
quite low and the default value $\beta=0.5$ is often
a good choice for training.
Further, supervision for only part of the training data $\vec x^i$
is obviously possible.

\subsection*{Experiments}
We demonstrate the behavior of median NG for a variety of
biomedical benchmark problems.
NG is intended for unsupervised data inspection, i.e.\
it can give hints on characteristic clusters and neighborhood
relationships in large data sets.
For the benchmark examples, prior class information is available,
such that we can evaluate the methods by means
of their classification accuracy.
For semisupervised learning, prototypes
and corresponding class labels are 
directly obtained from the learning algorithm. For unsupervised
training, posterior labeling of the prototypes based on
a majority vote of their receptive fields can be used.
For all experiments, repeated cross-validation has been used for
the evaluation.

\subsubsection*{Wisconsin breast cancer}
The Wisconsin breast cancer diagnostic database is a standard benchmark set
from clinical proteomics \cite{mangasarian}.
It consists of
569 data points described by 30 real-valued input features:
digitized images of a fine needle aspirate of breast mass are
described by characteristics such as form and texture
of the cell nuclei present in the image.
Data are labeled by two classes, benign and malignant.

The data set is contained in the Euclidean space such that
we can compare all clustering versions as introduced above
for this data set using the Euclidean metric.
We train 40 neurons using 200 epochs.
The dataset is standardized to z-scores and randomly split into two halfs for each run.
The result on the test set averaged over
100 runs is reported.
We obtain a test set accuracy of
$0.957$ for the supervised version
and $0.935$ for the unsupervised version, both setting
$\beta=0.1$ which is optimum for these
cases.
Results for simple K-means without neighborhood cooperation
yield an accuracy 0.938
for standard (unsupervised) K-means
resp.\ 0.941 for supervised K-means.
Obviously, there are only minor, though
significant differences of the results of
the different clustering variants on this data
set:
incorporation of neighborhood cooperation allows to improve K-means,
incorporation of label information
allows to improve fully unsupervised clustering.
As expected, Euclidean clustering is superior to median versions
(using the squared Euclidean norm) because 
the number of possible prototype locations is
reduced for median clustering.
However, the difference is only $1.3\%$,
which is quite remarkable because of the
comparably small data set, thus
dramatically reduced flexibility of prototype locations.

The article \cite{mangasarian} reports a test set accuracy of
0.97\% using 10-fold cross-validation and a supervised
learning algorithm
(a large margin linear classifier including feature selection).
This differs from our best classification result by 1.8\%. 
Thereby, the goal of our approach is a faithful prototype-based representation
of data, such that the result is remarkable.

\subsubsection*{Chromosomes}
The Copenhagen chromosomes database is a benchmark from
cytogenetics \cite{lundsteen}.
A set of 4200 human chromosomes from 22 classes (the autosomal chromosomes)
are represented by the grey levels of their images and transferred to
strings which represent the profile of
the chromosome by the thickness of their silhouettes.
This data set is non-Euclidean,
consisting of strings of different length,
and standard neural clustering cannot be used.
Median versions, however, are directly applicable.
The edit distance (also known as the Levenshtein distance
\cite{Levenshtein1966}), is a typical distance measure for two strings
of different length,
as described in \cite{juan,neuhaus}.
In our application,
distances of two strings are computed using the
standard edit distance whereby substitution costs are given by the
signed difference of the entries and insertion/deletion costs
are given by 4.5 \cite{neuhaus}.

The algorithms have been run using 100 neurons and 100 epochs per run.
Supervised median neural gas achieves an accuracy
of 0.89 for $\beta=0.1$.
This improves by 6\% compared to median K-means.
A larger number of prototypes allows to further improve this result:
500 neurons yield an accuracy of
0.93 for supervised median neural gas clustering
and 0.91 for supervised median K-means clustering,
both taken for $\beta=0.1$.
This is already close to the results of fully supervised
k-nearest neighbor classification
which uses all points of the training set for classification.
12-nearest neighbors with the standard edit distance
yields an accuracy 0.944 as reported in \cite{juan}
whereas more compact classifiers such as feedforward networks
or hidden Markov models only achieve an accuracy
less than 0.91, quite close to our results for only
100 prototypes.

\subsection*{Proteins}
The evolutionary distance of 226 globin proteins is
determined by alignment as described in \cite{vingron}.
These samples originate from different
protein families:
hemoglobin-$\alpha$, hemoglobin-$\beta$,
myoglobin, etc.
Here, we distinguish five classes
as proposed in \cite{haasdonk}:
HA, HB, MY, GG/GP, and others.

\begin{figure}[tb]
\begin{center}
\includegraphics[width=\textwidth]{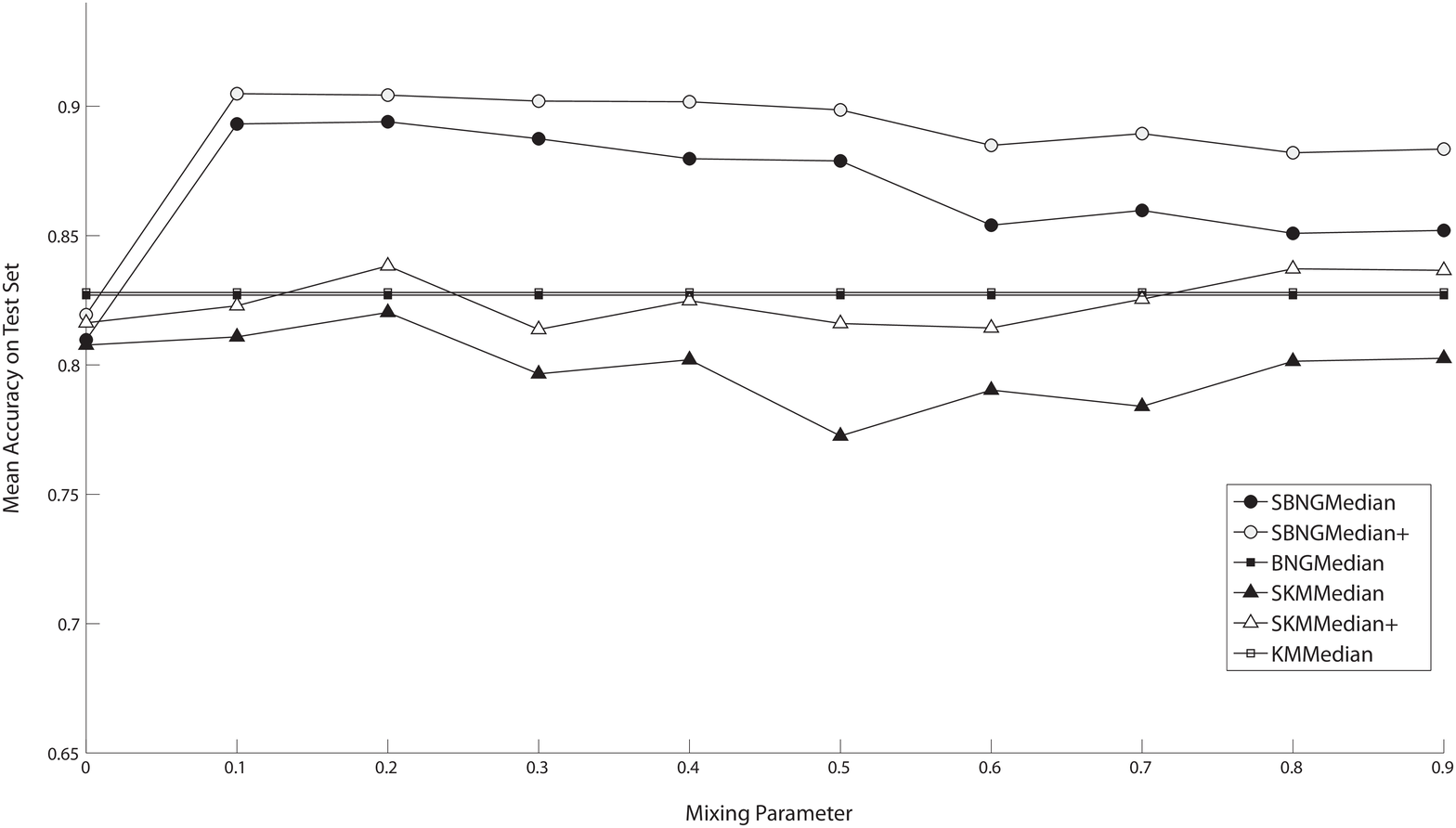}
\end{center}
\caption{Results of the methods for the protein database
using alignment distance
and varying mixing parameter $\beta$.
The version indicated with $+$ refers to (optimum) posterior labeling of prototypes.}
\label{proteins}
\end{figure}

We use 30 neurons and 300 epochs per run.
The accuracy on the test set averaged over
50 runs is reported in Fig.~\ref{proteins}.
Here, optimum mixing parameters can be observed for
supervised median neural gas and
$\beta\in[0.5,0.9]$, indicating
that the statistics of the inputs guides the way
towards a good classification accuracy.
However, an integration of the labels
improves the accuracy by nearly 10\% compared to 
fully unsupervised clustering.
As beforehand, integration of neighborhood cooperation
is well suited in this scenario.
Unlike the results reported in \cite{haasdonk} for
SVM which uses one-versus-rest encoding, the classification
in our setting is given by only one clustering model.
Depending on the choice of the kernel,
\cite{haasdonk} reports errors which approximately add up
to 0.04 for the leave-one-out error. 
This result, however, is not comparable to our
results due to the different error measure.
A 1-nearest neighbor classifier yields an accuracy
0.91 for our setting (k-nearest neighbor for larger k is worse; \cite{haasdonk}
reports an accumulated leave-one-out error of
0.065 for 1-nearest neighbor) which
is comparable to our (clustering) results.

Thus, unsupervised or semi-supervised data
inspection which accounts for both, data statistics and 
prior labeling, reaches a classification accuracy 
comparable to fully supervised approaches, i.e.\
the clusters found by median NG are meaningful in these cases.

\section{Fast implementations}\label{sectionFastMedianClustering}
As pointed out previously, the computational cost of median clustering
algorithms is quite high: the exhaustive search for the best prototypes leads
to a cost of $\mathcal{O}(N^2K)$ per epoch (for both median NG and median
SOM). This cost is induced by the need for evaluating a sum of the following
form
\[
\sum_{i=1}^N\alpha_{i,l,j}d(\vec x^i,\vec x^l),
\]
where $\vec x^l$ is a candidate for prototype $j$. Evaluating this sum is a
$\mathcal{O}(N)$ operation which has to be repeated for each candidate ($N$
possibilities) and for each prototype ($K$ times). As the coefficients of the
sums depends on $j$, it might seem at first glance that there is no way to
reduce the cost. 

\subsection{Block summing}
However, the prior structure of the SOM can be leveraged to reduce the total
cost to $\mathcal{O}(N^2+NK^2)$ per epoch \cite{FastDSOM}. Let $C^*_j$ denote
the receptive field of prototype $j$, more precisely 
\[
C^*_j:=\left\{i\in\{1,\ldots,N\}\mid I^*(\vec x^i)=j\right\}.
\]
Then the prototype $\vec w^j$ is given by
\[
\vec w^j=\arg\min_{\vec x^l}\sum_{k=1}^Kh_\sigma(\nd(k,j))\sum_{i\in
  C^*_k}d(\vec x^i,\vec x^l). 
\]
The main advantage of this formulation over the standard one is that there is
now a clean separation between components that depend on $j$ (the
$h_\sigma(\nd(k,j))$ terms) and those that do not (the sums $\sum_{i\in
  C^*_k}d(\vec x^i,\vec x^l)$). This leads to the following version of the
median SOM \cite{FastDSOM}:
\begin{center}
\begin{tabular}{l}
init\\
repeat\\
\hspace*{1em} $C^*_j:=\left\{i\in\{1,\ldots,N\}\mid I^*(\vec x^i)=j\right\}$ (\emph{receptive field calculation})\\
\hspace*{1em} $S(k,l):=\sum_{i\in
  C^*_k}d(\vec x^i,\vec x^l)$ (\emph{block summing})\\
\hspace*{1em} $\vec w^j:=\arg\min_{\vec
  x^l}\sum_{k=1}^Kh_\sigma(\nd(k,j))S(k,l)$ (\emph{prototype calculation})
\end{tabular}
\end{center}
There are $N\times K$ block sums $S(k,l)$ which can be computed in
$\mathcal{O}(N^2)$ operations as the $(C^*_k)_{1\leq k \leq N}$ form a
partition of the dataset. Then the exhaustive search involves only summing $K$
values per candidate prototype (and per prototype), leading to a total cost of
$\mathcal{O}(NK^2)$ (of the same order as the computation of the receptive
fields). The total computational load is therefore $\mathcal{O}(N^2+NK^2)$. In
practice, the speed up is very high. For instance, with the optimized Java
implementation proposed in \cite{FastDSOM}\footnote{Available at
  \url{http://gforge.inria.fr/projects/somlib/}}, a standard
$\mathcal{O}(N^2K)$ implementation of the median SOM uses approximately $5.7$
seconds per epoch on the Chromosomes dataset ($N=4200$) for $K=100$
prototypes (arranged on a $10\times 10$ hexagonal grid) on a standard
workstation\footnote{AMD Athlon 64 3000+ processor 
  with 2GB of main memory, running Fedora Linux 7 and with Sun 1.6 java
  virtual machine in server mode}. Under identical conditions, the above
algorithm uses only $0.25$ second per epoch while providing exactly the same
results (this is 23 times faster than the standard implementation, see Table \ref{table:SOM} for a summary of the timing results for the Median SOM variants). 

Obviously, the speed up strongly depends on both $N$ and $K$. For example if
$K$ is raised to $484$ (for a regular hexagonal grid of size $22\times 22$),
the standard implementation of the median SOM uses approximately $30.1$
seconds per epoch, with the block summing algorithm uses only $3.67$ seconds
per epoch. This is still more than 8 times faster than the standard
implementation, but as expected the speed up factor is worse than with a lower
value of $K$. Nevertheless as reflected in the theoretical cost analysis,
extensive simulations conducted in \cite{FastDSOM} have shown that the block
summing algorithm is always faster than the standard approach which has
therefore no reason to be used.

\subsection{Heuristic search}
Additional reduction in the actual running time of the median SOM
can be obtained via the branch and bound principle from combinatorial
optimization \cite{ConanGuezRossiIWANN2007}. The goal of branch and bound
\cite{LandDoig1960} is to avoid to perform an exhaustive search to solve a
minimization problem by means of two helper procedures. The first procedure is
a partition method to be applied to the search space (the \emph{branch} part
of the method). The second procedure provides \emph{quickly} a guaranteed
lower bound of the criterion to be minimized on any class of the partition of
the search space (the \emph{bound} part of the method). By \emph{quickly} one
means faster than an exhaustive evaluation of the criterion on the class. 

A standard implementation of the minimization of a function $f$ by an
exhaustive search on the search space $\mathcal{S}$ proceeds as follows:
\begin{center}
\begin{tabular}{l}
initialise $\mathit{best}$ to $s_1\in \mathcal{S}$\\
initialise $\mathit{fbest}$ to $f(\mathit{best})$\\
for all $s\in \mathcal{S}\setminus\{s_1\}$ do \\
\hspace*{1em} compute $f(s)$\\
\hspace*{1em} if $f(s)<\mathit{fbest}$ update $\mathit{best}$ to $s$ and $\mathit{fbest}$ to $f(s)$
\end{tabular}
\end{center}
To save some evaluations of $f$, a branch and bound search proceeds as follows:
\begin{center}
\begin{tabular}{l}
compute $C_1,\ldots,C_K$ a partition of $\mathcal{S}$\\
initialise $\mathit{fbest}$ and $\mathit{best}$ by an exhaustive search in $C_1$\\
for $i=2,\ldots,K$ \\
\hspace*{1em} compute a lower bound $g$ for $f$ on $C_i$\\
\hspace*{1em} if $g<\mathit{fbest}$ update $\mathit{best}$ and $\mathit{fbest}$ by an exhaustive search in $C_i$\\
\end{tabular}
\end{center}
The gain comes from the possibility of pruning entire regions (classes) of the
search space when the lower bound of the criterion $f$ on such a region is
higher than the best value found so far. The best gain is achieved when all
regions except $C_1$ are pruned. Obviously, the order in which the partition
classes are searched is crucial in obtaining good performances. 

In the median SOM, the search space is the dataset. This provides a natural
branching procedure as the receptive fields $(C^*_j)_{1\leq j\leq K}$ of the prototypes of the SOM build
a partition of the dataset. If the receptive fields have comparable sizes,
i.e., around $N/K$, branch and bound can reduce the search space for each
prototype from a size of $N$ to a size of $N/K$, in the optimal case (perfect
branching). This could reduce the cost of the search from $\mathcal{O}(NK^2)$
to $\mathcal{O}(NK+K^2)$. Indeed, in the ideal case, one would only evaluate
the quality criterion $\sum_{k=1}^Kh_\sigma(\nd(k,j))S(k,l)$ for candidate
prototypes from one cluster (this would cost
$\mathcal{O}(K(N/K))=\mathcal{O}(N)$) and then compare the best value to the
lower bound of each other cluster ($\mathcal{O}(K)$ additional operations).

The bounding procedure needs to provide a tight lower bound for the following
quantity 
\[
\min_{\vec x^l\in C^*_m}\sum_{k=1}^Kh_\sigma(\nd(k,j))S(k,l).
\]
A class of lower bounds is given by the following equation
\begin{equation}
  \label{eq:BB:LowerBounds}
\eta(m,j,\Theta):=\sum_{k\in\Theta}h_\sigma(\nd(k,j))\min_{\vec x^l\in C^*_m}S(k,l),
\end{equation}
where $\Theta$ is a subset of $\{1,\ldots,K\}$. There are several reasons for
using such bounds. First the quantity $\min_{\vec x^l\in C^*_m}S(k,l)$ depends
only on $k$ and $l$: it can be computed once and for all before the
exhaustive search phase (in fact in parallel with the computation of the block
sum $S(k,l)$ itself). The additional cost is negligible compared to other
costs (there are $K^2$ values which are computed in $\mathcal{O}(NK)$
operations). However, computing $\eta(m,j,\{1,\ldots,K\})$ is costly, as the
search process will need this bound for all $m$ and $j$, leading to a total
cost of $\mathcal{O}(K^3)$: this is small compared to $\mathcal{O}(N^2+NK^2)$
but not negligible when $K$ is large, especially compared to the best case
cost ($\mathcal{O}(N^2+NK+K^2)$ with perfect branching).

It is therefore interesting in theory to consider strict subsets of
$\{1,\ldots,K\}$, in particular the singleton $\Theta=\{j\}$ which leads to
the very conservative lower bound $h_\sigma(\nd(j,j))\min_{\vec x^l\in
  C^*_m}S(j,l)$, for which the computation cost is only $\mathcal{O}(K^2)$ for
all $m$ and $j$. Despite its simplicity, this bound leads to very good results
in practice \cite{ConanGuezRossiIWANN2007} because when the neighborhood
influence is annealed during training, $h_\sigma(\nd(k,j))$ gets closer and
closer to the Kronecker function $\delta_{k,j}$.

Compared to the improvements generated by reducing the complexity to
$\mathcal{O}(N^2+NK^2)$, the speed up provided by branch and bound is
small. Under exactly the same conditions as in the previous section, the time
needed per epoch is $0.22$ second (compared to $0.25$) when the bounds are
computed with $\Theta=\{j\}$ and $0.14$ second when $\Theta=\{1,\ldots,K\}$
(which shows that perfect branching does not happen as the $\mathcal{O}(K^3)$
cost of the bounds calculation does not prevent from getting a
reasonable speed up). Complex additional programming tricks exposed in
\cite{ConanGuezRossiIWANN2007,FastDSOM} can reduce even further the running
time in some situation (e.g., when $K$ is very large), but on the Chromosomes
dataset with $K=100$, the best time is obtained with the algorithm described
above. The speed up compared to a naive implementation is nevertheless quite
large as the training time is divided by 40, while the results are guaranteed
to be exactly identical. 

\begin{table}[htbp]
  \centering
  \begin{tabular}{lcc}
Algorithm & $K=100$ & $K=484$ \\\hline
Standard implementation &$5.74$ &$30.1$\\
Block Summing           &$0.248$&$3.67$\\
Branch and bound $\Theta=\{j\}$ & $0.224$& $2.92$\\
Branch and bound $\Theta=\{1,\ldots,K\}$ & $0.143$& $1.23$\\
Branch and bound $\Theta=\{1,\ldots,K\}$ and early stopping & $0.141$& $0.933$\\\hline
  \end{tabular}
  \caption{Average time needed to complete an epoch of the Median SOM
    (in seconds) for the Chromosomes dataset}
  \label{table:SOM}
\end{table}

When $K$ is increased to $484$ as in the previous section, the time per epoch
for $\Theta=\{j\}$ is $2.92$ seconds and $1.23$ seconds when
$\Theta=\{1,\ldots,K\}$. In this case, the ``early stopping'' trick described
in \cite{ConanGuezRossiIWANN2007,FastDSOM} (see also the next Section) can be
used to bring down this time to $0.93$ second per epoch. This is more than 32
times faster than the naive implementation and also almost 4 times faster than
the block summing method. Branch and bound together with early stopping
complement therefore the block summing improvement in the following sense:
when $K$ is small (of the order of $\sqrt{N}$ or below), block summing is
enough to reduce the algorithmic cost to an acceptable
$\mathcal{O}(N^2)$ cost. While $K$ grows above this limit, branch and bound and
early stopping manage to reduce the influence of the $\mathcal{O}(NK^2)$ term on the total running time. 

\subsection{Median Neural Gas}
Unfortunately the solutions described above cannot be applied to
median NG, as there is no way to factor the computation of
$\sum_{i=1}^Nh_{\sigma}(k_{ij})d(\vec x^i,\vec x^l)$ in a similar way as the
one used by the median SOM. Indeed, the idea is to find in the sum sub-parts
that do not depend on $j$ (the prototype) so as to re-use them for all the
prototypes. The only way to achieve this goal is to use a partition on the
dataset (i.e., on the index $i$) such that the values that depend on $j$,
$h_{\sigma}(k_{ij})$, remain constant on each class. This leads to the
introduction of a partition $R^j$ whose classes are defined by
\[
R^j_k=\left\{i\in\{1,\ldots,N\}\mid\rk(\vec x^i,\vec w^j)=k\right\}.
\]
This is in fact the partition induced by the equivalence relation on
$\{1,\ldots,N\}$ defined by $i\sim_j i'$ if and only if
$h_{\sigma}(k_{ij})=h_{\sigma}(k_{i'j})$. Using this partition, we have
\begin{equation}\label{eq:ng:factorized}
\sum_{i=1}^Nh_{\sigma}(k_{ij})d(\vec x^i,\vec
x^l)=\sum_{k=1}^Kh_{\sigma}(k)\sum_{i\in R^j_k}d(\vec x^i,\vec x^l).
\end{equation}
At first glance, this might look identical to the factorisation used for the
median SOM. However there is a crucial difference: here the partition
\textbf{depends} on $j$ and therefore the block sums $\sum_{i\in R^j_k}d(\vec
x^i,\vec x^l)$ cannot be precomputed (this factorization will nevertheless
prove very useful for early stopping). 

For both algorithms, the fast decrease of $h_\sigma$
suggests an approximation in which small values of $\alpha_{i,l,j}$ (i.e., of
$h_{\sigma}(k_{ij})$ and of $h_\sigma(\nd(k,j))$) are discarded. After a few
epochs, this saves a lot of calculation but the cost of initial epochs remains
unchanged. Moreover, this approximation scheme changes the results of the
algorithms, whereas the present section focuses on exact and yet fast
implementation of the median methods.

A possible source of optimization for median NG lies in the so called ``early
stopping'' strategy exposed for the median SOM in
\cite{ConanGuezRossiIWANN2007,FastDSOM}. The idea is to leverage the fact that
the criterion to minimize is obtained by summing positive values. If the sum
is arranged in such as way that large values are added first, the partial
result (i.e., the sum of the first terms of the criterion) can exceed the best
value obtained so far (from another candidate). Then the loop that implements
the summing can be stopped prematurely reducing the cost of the evaluation of
the criterion. Intuitively, this technique works if the calculation are
correctly ordered: individual values in the sum should be processed in
decreasing order while candidate prototypes should be tested in order of
decreasing quality. 

For the median SOM, as shown in
\cite{ConanGuezRossiIWANN2007,FastDSOM}, early stopping, while interesting in
theory, provides speed up only when $K$ is large as the use of block summing
has already reduced the cost of the criterion evaluation from $\mathcal{O}(N)$
to $\mathcal{O}(K)$. On the Chromosomes dataset for instance, the previous
Section showed that early stopping gains nothing for $K=100$ and saves
approximately $24\%$ of the running time for $K=484$ (see Table \ref{table:SOM}).  

However, as there is no simplification in the criterion for median NG, early
stopping could cause some improvement, especially as the sorting needed to
compute the $\rk$ function suggests an evaluation order for the criterion and
an exploration order during the exhaustive search. 

The computation of $\vec w^j:=\arg\min_{x^l}\sum_{i=1}^Nh_{\sigma}(k_{ij})d(\vec x^i,\vec 
x^l)$ by an early stopping algorithm takes the following generic form:
\begin{center}
\begin{tabular}{l}
$q:=\infty$\\
for $l\in\{1,\ldots,N\}$ \emph{(candidate loop)}\\
\hspace*{1em} $s:=0$\\
\hspace*{1em} for $i\in\{1,\ldots,N\}$ \emph{(inner loop)}\\
    \hspace*{2em} $s:=s+h_{\sigma}(k_{ij})d(\vec x^i,\vec x^l)$\\
    \hspace*{2em} if $s>q$ break the inner loop\\
\hspace*{1em} endfor \emph{(inner loop)}\\
\hspace*{1em} if $s<q$ then\\
    \hspace*{2em} $q:=s$\\
    \hspace*{2em} $\vec w^j:=\vec x^l$\\
\hspace*{1em} endif\\
endfor \emph{(candidate loop)}
\end{tabular}
\end{center}

\subsubsection{Ordering}
Both loops (\emph{candidate} and \emph{inner}) can be performed in specific
orders. The \emph{candidate loop} should analyse the data points $\vec x^l$ in
decreasing quality order, i.e., it should start with $\vec x^l$ such that
$\sum_{i=1}^Nh_{\sigma}(k_{ij})d(\vec x^i,\vec x^l)$ is small and end with the
points that have a large value of this criterion. This optimal order is
obviously out of reach because computing it implies the evaluation of all the
values of the criterion, precisely what we want to avoid. However, Median
Neural Gas produces clusters of similar objects, therefore the best candidates
for prototype $\vec w^j$ are likely to belong to the receptive field of this
prototype at the previous epoch. A natural ordering consists therefore in
trying first the elements of this receptive field and then elements of the
receptive fields of other prototypes in increasing order of dissimilarities
between prototypes. More precisely, with
\[
C_k:=\left\{i\in\{1,\ldots,N\}\mid I(\vec x^i)=k\right\},
\]
the \emph{candidate loop} starts with $C_{j_1}=C_j$ and proceeds through the
$(C_{j_i})_{2\leq i\leq K}$ with $d(\vec w^j,\vec w^{j_{i-1}})\leq d(\vec w^j,\vec
w^{j_i})$. Computing this order is fast ($\mathcal{O}(K^2\log K)$ operations
for the complete prototype calculation step). 

Another solution consists in ordering the \emph{candidate loop} according to
increasing ranks $k_{lj}$ using the partition $R^j$ defined previously instead
of the receptive fields. 

The \emph{inner loop} should be ordered such that $s$ increases as quickly as
possible: high values of $h_{\sigma}(k_{ij})d(\vec x^i,\vec x^l)$ should be
added first. As for the \emph{candidate loop} this cannot be achieved exactly
without loosing all benefits of the early stopping. A simple solution consists
again in using the partition $R^j$. The rational is that the value of
$h_{\sigma}(k_{ij})d(\vec x^i,\vec x^l)$ is likely to be dominated by the
topological term $h_{\sigma}(k_{ij})$, especially after a few epochs when
$h_{\sigma}$ becomes peaked in $0$. This ordering has three additional
benefits. As explained above, the factorized representation provided by
equation \eqref{eq:ng:factorized} allows to save some calculations. Moreover,
$R^j$ can be computed very efficiently, as a side effect of computing the
ranks $k_{ij}$. If they are stored in a $NK$ array, $R^j$ is obtained by a
single pass on the index set $\{1,\ldots,N\}$. Computing the $R^j$ for all $j$
has therefore a $\mathcal{O}(NK)$ cost. Finally, $R^j$ can also be used for
ordering the \emph{candidate loop}. 

It should be noted that using different orderings for the \emph{candidate
  loop} can lead to different final results for the algorithm in case of ties
between candidates for prototypes. In practice, the influence of such
differences is extremely small, but contrarily to the SOM for which all
experiments produced in \cite{ConanGuezRossiIWANN2007,FastDSOM} and in this
chapter gave exactly the same results, regardless of the actual
implementation, variants of the Median Neural Gas exhibit a small variability
in the results (less than one percent of differences in the quantization error,
for instance). Those differences have been neglected as they could be
suppressed via a slightly more complex implementation in which ties between
candidate prototypes are broken at random (rather than via the ordering);
using the same random generator would produce exactly the same results in all
implementations. 

\subsubsection{Early stopping granularity}
Experiments conducted in \cite{ConanGuezRossiIWANN2007,FastDSOM} have shown
that early stopping introduces a non negligible overhead to the \emph{inner
  loop} simply because it is the most intensive part of the algorithm which is
executed $N^2K$ times in the worst case. A coarse grain early stopping strategy
can be used to reduce the overhead at the price of a more complex code and of
less early stops. The idea is to replace the standard \emph{inner loop} by the
following version:
\begin{center}
\begin{tabular}{l}
$s:=0$\\
for $m\in\{1,\ldots,M\}$ \emph{(monitoring loop)}\\
\hspace*{1em} for $i\in B_m$ \emph{(internal loop)}\\
    \hspace*{2em} $s:=s+h_{\sigma}(k_{ij})d(\vec x^i,\vec x^l)$\\
\hspace*{1em} endfor \emph{(internal loop)}\\
\hspace*{1em} if $s>q$ break the monitoring loop\\
endfor \emph{(monitoring loop)}
\end{tabular}
\end{center}
The main device is a partition $B=(B_1,\ldots,B_M)$ of $\{1,\ldots,N\}$ which
is used to divide the computation into uninterrupted calculations
(\emph{internal loops}) and to check on a periodic basis by the \emph{monitoring
  loop}. The value of $M$ can be used to tune the grain of the early stopping
with a classical trade-off between the granularity of the monitoring and its
overhead. In this chapter, we have focused on a particular way of implementing
this idea: rather than using an arbitrary partition $B$, we used the $R^j$
partition. It has two advantages over an arbitrary one: is provides an
interesting ordering of the \emph{monitoring loop} (i.e., in decreasing order of
$h_{\sigma}(k_{ij})$) and allows the code to use the factorized equation
\eqref{eq:ng:factorized}. 

\begin{table}[htbp]
  \centering
  \begin{tabular}{lcc}
Algorithm & $K=100$ & $K=500$ \\\hline
Standard implementation &$5.31$&$25.9$\\
Early stopping without ordering &$5.26$&$24.3$\\
Early stopping with \emph{candidate loop} ordering &$4.38$&$20.7$\\
Full ordering and fine grain early stopping&$1.16$ & $7.45$\\
Full ordering and coarse grain early stopping&$0.966$&$5.91$\\\hline
  \end{tabular}
  \caption{Average time needed to complete an epoch of the Median Neural Gas
    (in seconds) for the Chromosomes dataset}
  \label{table:NeuralGas:EarlyStopping}
\end{table}

\subsubsection{Experiments}
Variants of the early stopping principle applied to Median Neural Gas were
tested on the Chromosomes ($N=4200$) with $K=100$ and $K=500$. They are
summarized in Table \ref{table:NeuralGas:EarlyStopping}. The standard
implementation corresponds to the exhaustive search $\mathcal{O}(N^2K)$
algorithm. The need for ordering is demonstrated by the results obtained by a
basic implementation of early stopping in which the natural data ordering is
used for both \emph{candidate} and \emph{inner} loops. Moreover, while the
\emph{candidate loop} order based on receptive fields reduces the running
time, the gain remains limited when the \emph{inner} loop is not ordered (the
running time is reduced by approximately 20\% compared to the standard
implementation). 

Much better improvements are reached when the $R^j$ partitions are used to
order both loops. The running time is divided by more than $5$ for $K=100$ and
by more than $4$ for $K=500$, when a coarse grain early stopping method is
used. The fine grain version is slightly less efficient because of the
increased overhead in the \emph{inner loop}. 

The structure of the Median Neural Gas algorithm prevents the use of the block
summing trick which is the main source of improvement for the Median SOM. In
the case of Neural Gas, early stopping provides better improvement over the
state-of-the-art implementation, than it does for the SOM, because it targets
an internal loop with $\mathcal{O}(N)$ complexity whereas the block summing
approach leads to a $\mathcal{O}(K)$ inner loop. In the end, the optimized
Median SOM remains much faster than the Median Neural Gas (by a factor
6). However, the SOM is also very sensitive to its initial configuration
whereas Neural Gas is rather immune to this problem. In practice, it is quite
common to restart the SOM several times from different initial configuration,
leading to quite comparable running time for both methods.

\section{Approximate patch clustering for large data sets}
A common challenge today \cite{challenge}, arising especially in computational biology,
image processing, and physics, are huge datasets whose pairwise dissimilarities
cannot be hold at once within random-access memory during computation,
due to the sheer amount of data (a standard workstation with 4 GB of
  main memory cannot hold more than $N=2^{15}$ data points when they are
  described by a symmetric dissimilarity matrix).
Thus, data access is costly and only a few, ideally at most one
pass through the data set is still affordable. 

Most work in this area can be found in the context of heuristic (possibly hierarchical)
clustering on the one side and classical K-means clustering on the other side.
Heuristic algorithms often directly assign data consecutively
to clusters based on the distance within the cluster and allocate new clusters
as required. Several popular methods include CURE, STING, and BIRCH \cite{cure,sting,birch}.
These methods do not rely on a cost
function such that an 
incorporation of label information into the clustering becomes difficult.

Extensions of K-means clustering can be distinguished into methods
which provide guarantees on the maximum difference of the result
from classical K-means, such as presented in the approaches
\cite{guha,jin}. However, these variants use resources
which scale in the worst case with a factor depending on $N$
($N$ being the number of points)
with respect to memory requirements or passes through the data set.
Alternatives are offered by variants of K-means which do not provide approximation
guarantees, but which can be strictly limited with respect to space requirements
and time. An early approach has been proposed in
\cite{bradley}:
data are clustered consecutively in small patches, whereby
the characteristics of the data and the possibility to compress subsets of data are
taken into account. A simpler although almost as efficient method has
been proposed in \cite{farnstrom}: Standard K-means is performed
consecutively for patches of the data whereby each new
patch is enriched by the prototypes obtained in the previous
patch. A sufficient statistics of the outcome of the last run can
thereby easily be updated in a consecutive way, such that
the algorithm provides cluster centres after only one pass through
the data set, thereby processing the data consecutively in patches of
predefined fixed size.

Some of these ideas have been transferred to topographic maps:
the original median SOM \cite{mediansom} proposes simple sampling to
achieve efficient results for huge data sets. Simple sampling is
not guaranteed to preserve the statistics of the data and
some data points might not be used for training at all, because
of which reason \cite{mediansom} proposes to use all data in the last
run.
An approach which uses all available statistics consists in
an extension of patch clustering towards neural gas and
alternatives, as proposed in \cite{AleHamKla07,esann}.
This method processes data in patches, thereby integrating the sufficient
statistics of results of the previous run, such that all
available information processed so far is used in each
consecutive clustering step. Since the runs rely on a statistics
of the data, the overall result only approximates the optimum
solution obtained by standard batch clustering.
However, in practice, results are quite good.

An extension of this method
to median clustering is possible and yields promising results, as proposed
in \cite{aipr}.
Here we give an introduction to this simple and powerful
extension of median clustering to huge data sets. 
As a side effect, this method dramatically reduces
the complexity of clustering to linear complexity in $N$,
$N$ being the number of data points.

Assume, as before,
a dissimilarity matrix $D$ with entries $d_{ij}$
representing the
dissimilarity of patterns.
Here we assume symmetry of the matrix, but no further requirements
need to be fulfilled.
For huge data sets, only parts of the matrix $D$ fit into main memory.
We assume that access to single elements of the matrix is possible
at any time, e.g.\ the elements are stored in a database or the dissimilarity
values are computed on the fly by means of some
(possibly complex) dissimilarity measure (such as pairwise alignment
of proteins using FASTA).
The exact way how dissimilarities are accessed is
not relevant for patch clustering.

During processing of patch Median NG, $n_p$ disjoint patches of fixed size $p = \lfloor m / n_p \rfloor$
are taken from the dissimilarity matrix $D$ consecutively,\footnote{The remainder is no further
considered here for simplicity. In the practical
implementation the remaining datapoints are simply distributed
over the first $(M - p\cdot n_p)$ patches.}
where every patch
  $$P_i = \left( d_{st} \right)_{s,t = (i-1)\cdot p, \ldots, i\cdot p - 1} \:\in\:\mathbb{R}^{p \times p}$$
is a submatrix of $D$, representing
data points $(i-1)\cdot p$
to $i\cdot p - 1$.
The patches are small such that they fit into main memory.
The idea of the patch scheme is to add the prototypes from
the processing of the former patch $P_{i-1}$ as additional datapoints to
the current patch $P_i$, forming an extended patch $P^*_i$ to work on further.
The additional datapoints -- the former prototypes -- are weighted
according to the size of their receptive fields, i.e.\ how many
datapoints they have been representing in the former patch.
Therefore, every datapoint $\vec x^i$, as a potential prototype, is equipped
with a multiplicity $m_i$, that is at first initialized with $m_i = 1$.
Unlike simple sampling strategies,
every point of the dataset is considered
exactly once
and a sufficient statistics of all already processed data
is passed to further patches by means of the weighted
prototypes.

Unlike the situation of patch NG in Euclidean space \cite{AleHamKla07,esann},
where inter-patch distances can always be recalculated with
help of the Euclidean metric,
we are now dealing with an unknown mathematical space.
We have to construct the extended patch from given dissimilarity data.
The extended patch $P^*_i$
is defined as

 \[  P^*_i = \left(
 \begin{array}{c@{\:}c@{\:}c@{\:}|@{\:}c@{\:}c@{\:}c@{\:}}
     & & & & & \\
     & d(N_{i-1}) & & \quad d(N_{i-1}, P_i)\quad & & \\
     & & & & & \\\hline
     & & & & & \\
     & & & & & \\
     & d(N_{i-1}, P_i)^T & & P_i & & \\
     & & & & & \\
     & & & & & \\
 \end{array}
 \right)
 \]

where $$ \begin{array}{l}

    d(N_{i-1}) = \left( d_{st} \right)_{s,t \in N_{i-1}} \:\in\:\mathbb{R}^{K \times K} \\[1ex]

    d(N_{i-1}, P_i) = \left( d_{st} \right)_{s \in N_{i-1}, t = (i-1)\cdot p, \ldots, i\cdot p - 1} \:\in\:\mathbb{R}^{K \times p}

 \end{array}$$
denote the inter-distances of former prototypes and the distances between former prototypes and current patch points, respectively.
Every point is weighted with a multiplicity $m_j$ which
is set to $1$ for all new points $j\in [(i-1)\cdot p,i\cdot p -1]$.
For points which stem from prototypes,
the multiplicity is set to the sum of the multiplicities
of all points in its receptive field.

To apply median clustering,
we have to incorporate 
these multiplicities into the learning scheme.
The cost function becomes
\begin{equation}
\hat E_{\mathrm{NG}}\sim \frac{1}{2N}\cdot \sum_{i=1}^N\sum_{j=1}^Kh_{\sigma}(\rk(\vec x^i,\vec w^j))\cdot m_j\cdot d(\vec x^i,\vec w^j)
\end{equation}
where, as before, prototype locations $\vec w^j$ are restricted
to data points.
Optimum ranks are obtained as beforehand.
Optimum prototypes are determined by means of the formula
\begin{equation}
\vec w^j=\argmin_{\vec x^l}\sum_{i=1}^Nh_{\sigma}(r_{ij})\cdot m_j\cdot d(\vec x^i,\vec x^l)
\end{equation}
Picking up the pieces, we obtain the following algorithm:\bigskip
\begin{center}
\begin{tabular}{l}
\noindent
\textbf{Patch Median Neural Gas}\\[1ex]

Cut the first Patch $P_1$\\
Apply Median NG on $P_1$ $\longrightarrow$ Prototypes $N_1$\\
Update Multiplicities $m_j$\\[1ex]

Repeat for $i = 2, \ldots, n_p$\\
\mbox{}\hspace*{1em} Cut patch $P_i$\\
\mbox{}\hspace*{1em} Construct Extended Patch $P^*_i$ using $P_i$ and $N_{i-1}$\\
\mbox{}\hspace*{1em} Apply modified Median NG with Multiplicities\\
\mbox{}\hspace*{2em} $\longrightarrow$ Prototypes $N_i$\\
\mbox{}\hspace*{1em} Update Multiplicities $m_j$\\[1ex]

Return final Prototypes $N_{n_p}$
\end{tabular}
\end{center}
Median SOM
can be extended
to patch clustering in a similar way.

We demonstrate the behavior of patch clustering on the breast
cancer data set which has been used beforehand.
Here, we compare data points with the Cosine
Measure
 $$d_{cos}(\vec x^i,\vec x^j) = 1 - \frac{\vec x^i\cdot \vec x^j}{\|\vec x^i\|_2\cdot\|\vec x^j\|_2}.$$
Standard median batch NG for 40 neurons and 100 epochs
yields an average classification accuracy of 0.95 in a repeated 10-fold
cross-validation.
In comparison, patch median NG with 5 patches, i.e.\ 114 data points per
patch, arrives at a classification accuracy of 0.94, yielding
only a slight decrease of the classification accuracy.

The complexity of standard median NG
is improved by incorporation of patches, as can be seen as follows.
Assume a fixed patch size $p$
independent of the number of datapoints, e.g.\
$p$ is chosen according to the main memory.
Then the algorithm uses only
${\cal O}(\frac{m}{p} \cdot (p+K)^2) = {\cal O}(m\cdot p + m\cdot K) = {\cal O}(m)$ entries
of the dissimilarity matrix, compared to ${\cal O}(m^2)$ in the original Median NG
method. Moreover, every epoch (within a patch) has complexity ${\cal O}(p^2)=\mbox{constant}$ as opposed to ${\cal O}(N^2)$ for an epoch in full median clustering.
Therefore the method does not only overcome the problem of limited memory,
it also dramatically accelerates the processing of datasets, what might be useful
in time critical applications.

\section{Discussion}
Neural clustering methods such as SOM and NG offer
robust and flexible tools for data inspection.  In biomedical domains,
data are often nonvectorial such that extensions of the original methods towards
general dissimilarity data have to be used. In this chapter, we
presented an overview about one particularly interesting technique which extends
NG and SOM towards dissimilarities by means of the generalized median.
Prototypes are restricted to data positions such that the standard cost functions 
are well defined and extensions such as supervision can easily be transferred to this setting.
Moreover, this way, clusters are represented in terms of typical exemplars from the data set, i.e.\
the idea offers a data representation which can be easily interpreted by experts in
biomedical domains.

These benefits are paid back by increased costs in a naive implementation of the
algorithms, the complexity of one epoch being of order $N^2$ instead of $N$,
where $N$ refers to the number of data points.
Since data are represented by a general $N\times N$ dissimilarity matrix instead of 
$N$ single vectors, these increased costs are to some extent unavoidable if 
the full information contained in the data is considered.
Nevertheless, a variety of structural aspects allow to reduce the costs
of median clustering in practical situations.

We discussed a variety of techniques which lead to a dramatic decrease of the training time 
while (approximately) preserving the quality of the original methods.
These approaches can be decomposed into exact methods which provably lead to the same
results as the original implementation and approximations which slightly reduce the
quality of the results in return for an improved efficiency.
Exact methods include
\begin{itemize}
\item block summing for median SOM due to the specific and fixed structure of the SOM
neighborhood; as pointed out in this chapter, block summing leads to a major reduction
of the computation time in this case. The method cannot be applied to NG, though, because 
NG does not rely on a priorly fixed lattice structure.
\item branch and bound methods which allow to reduce the number of necessary computations depending on
the situation at hand; usually, the computational savings strongly depend on the order in which
computations are performed. As pointed out in this chapter, branching can
be done with respect to candidate prototypes on the one hand and summands which contribute to the
overall cost associated with one prototype on the other hand.
For both settings, the topological ordering of the data suggests a natural decomposition
of the whole search space into parts. This procedure yields to a significant reduction
of the computational costs for NG in particular for later states of training with partially ordered setting. For SOM, the savings are only minor compared to savings by means of block summing,
though possibly significant depending on the number of prototypes.
\end{itemize}
These methods lead to the same results as a naive implementation but run in a fraction of the time.

Compared to these approaches, approximate methods
constitute a compromise of accuracy and complexity.
We presented a patch clustering approach for median clustering, which processes
data in patches of fixed size and integrates the results by means of the sufficient statistics
of earlier runs. This way, the computation time is reduced from ${\cal O}(N^2)$ to
${\cal O}(N)$. In particular, only a small part of the dissimilarity matrix
is considered in patch training. This has the additional benefit that, this way, only a finite and fixed memory size is required and the clustering method can readily be applied to huge streaming data
sets.
Further, since only a fraction of the dissimilarity matrix needs to be computed, this method 
is particularly suited for biomedical applications with complex dissimilarity measures
such as alignment distance.
\bibliographystyle{abbrv}
\bibliography{dagsofa3}
\end{document}